\documentclass[11pt,a4paper]{article}
\usepackage[hyperref]{acl2021}
\usepackage{times}
\usepackage{latexsym}

\usepackage{microtype}
\usepackage{multirow}
\usepackage{scalerel}
\usepackage{arydshln}
\usepackage{hyperref}
\usepackage{amsmath}
\aclfinalcopy 
\def\aclpaperid{2280} 

\newcommand{\refeqn}[1]{Equation \ref{#1}}
\newcommand{\reffig}[1]{Figure \ref{#1}}
\newcommand{\reftbl}[1]{Table \ref{#1}}
\newcommand{\refsec}[1]{Section \ref{#1}}
\newcommand{\refapp}[1]{Appendix \ref{#1}}

\newcommand{\method}{\textsc{MergeDistill}}

\graphicspath{ {./Figures/} }

\title{\method{}: Merging Pre-trained Language Models using Distillation}

\author{Simran Khanuja \\
    Google Research \\ India \\
  \texttt{simrankh@google.com} \\\And
 Melvin Johnson \\
  Google Research \\ USA \\
  \texttt{melvinp@google.com} \\ \And
  Partha Talukdar \\
  Google Research \\ India \\
  \texttt{partha@google.com}}
  
\date{}

\begin{document}
\maketitle
\begin{abstract}
Pre-trained multilingual language models (LMs) have achieved state-of-the-art results in cross-lingual transfer, but they often lead to an inequitable representation of languages due to limited capacity, skewed pre-training data, and sub-optimal vocabularies. This has prompted the creation of an ever-growing pre-trained model universe, where each model is trained on large amounts of language or domain specific data with a carefully curated, linguistically informed vocabulary. However, doing so brings us back full circle and prevents one from leveraging the benefits of multilinguality. To address the gaps at both ends of the spectrum, we propose \method{}, a framework to merge pre-trained LMs in a way that can best leverage their assets with minimal dependencies, using \emph{task-agnostic} knowledge distillation. We demonstrate the applicability of our framework in a practical setting by leveraging pre-existing teacher LMs and training student LMs that perform competitively with or even outperform teacher LMs trained on several orders of magnitude more data and with a fixed model capacity. We also highlight the importance of teacher selection and its impact on student model performance. 

\end{abstract}

\section{Introduction}

\begin{figure}
\begin{center}
\includegraphics[scale=0.19]{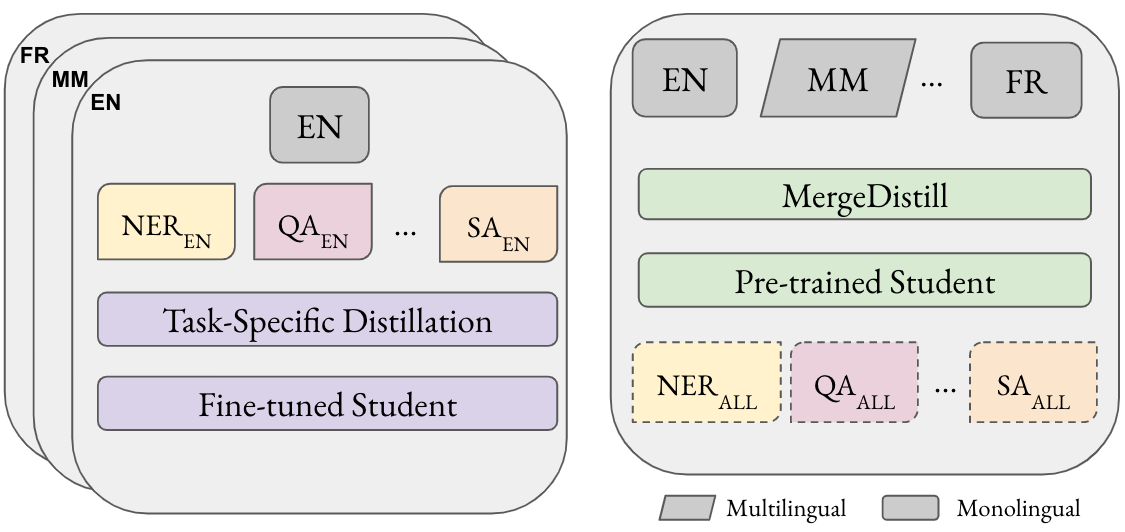} 
\caption{Previous works (left) typically focus on combining \emph{fine-tuned} models derived from a single pre-trained model using distillation. We propose \method{} to combine \emph{pre-trained} teacher LMs from multiple monolingual/multilingual LMs into a single \emph{multilingual task-agnostic} student LM.
}
\label{fig:distil_types}
\end{center}
\end{figure}

\begin{figure*}
\begin{center}
\includegraphics[scale=0.25]{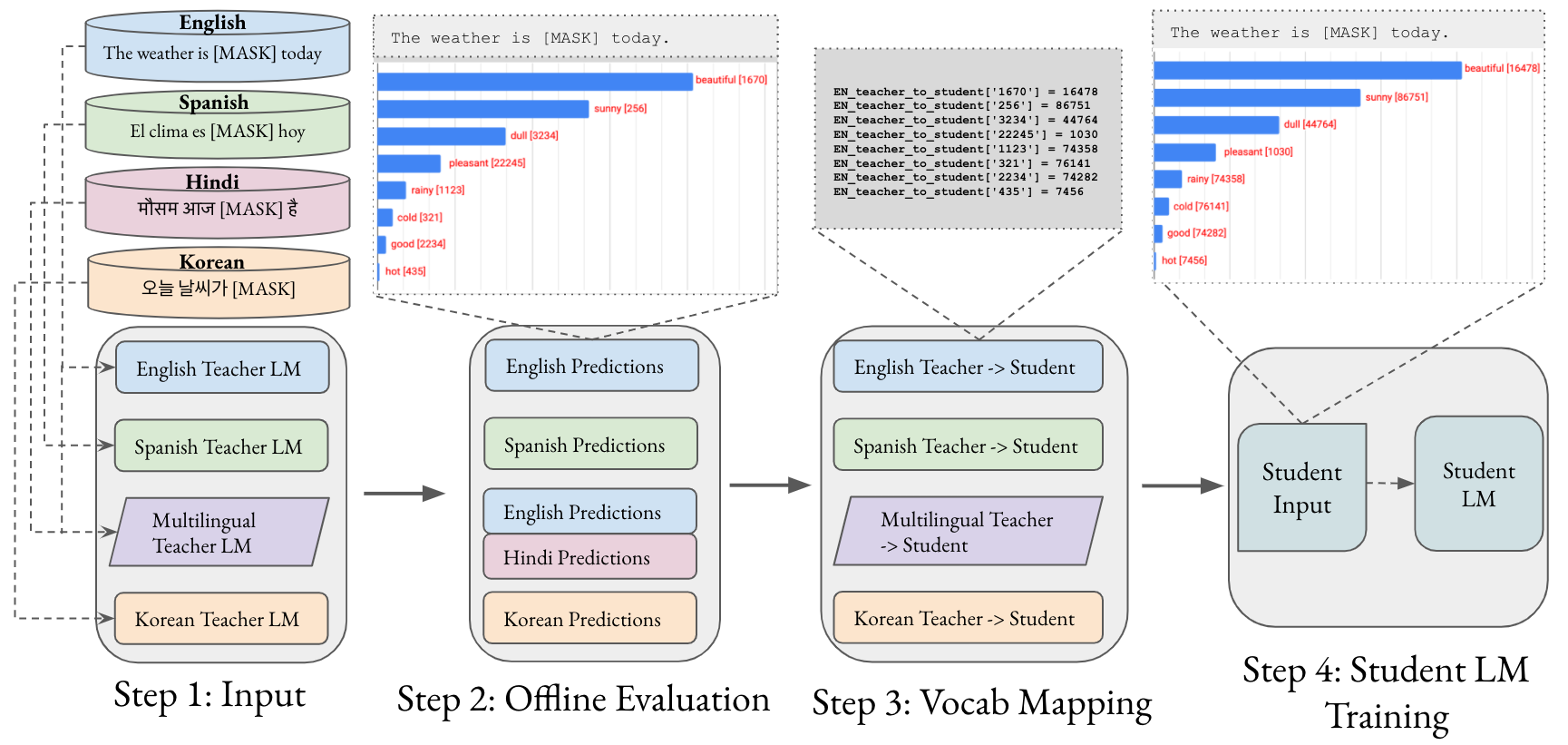} 
\caption{\textit{Overview of \method{}}: The input to \method{} is a set of pre-trained teacher LMs and pre-training transfer corpora for all the languages we wish to train our student LM on. Here, we combine four teacher LMs comprising of three monolingual (trained on English, Spanish and Korean respectively) and one multilingual LM (trained on English and Hindi). The student LM is trained on English, Spanish, Hindi and Korean. Pre-training transfer corpora for each language is tokenized and masked using their respective teacher LMs vocabulary. We then obtain predictions for each masked word in each language, by evaluating \emph{all} of their respective teacher LMs. For example, we evaluate English masked examples on both the monolingual and multilingual LM as shown. The student's vocabulary is a union of \emph{all} teacher vocabularies. Hence, the \emph{input, prediction and label indices} obtained from teacher evaluation are now mapped to the student vocabulary, and input to the student LM for training. Please refer to \refsec{workflow} for details.}
\label{fig:architecture}
\end{center}
\end{figure*}

While current state-of-the-art multilingual language models (LMs) \cite{devlin-etal-2019-bert, conneau-etal-2020-unsupervised} aim to represent  100+ languages in a single model, efforts towards building monolingual \cite{martin2019camembert, kuratov2019adaptation} or language-family based \cite{khanuja2021muril} models are only increasing with time \cite{rust2020good}. A single model is often incapable of effectively representing a diverse set of languages, evidence of which has been provided by works highlighting the importance of vocabulary curation and size \cite{chung-etal-2020-improving, artetxe-etal-2020-cross}, pre-training data volume \cite{liu-etal-2019-robust, conneau-etal-2020-unsupervised}, and the curse of multilinguality \cite{conneau-etal-2020-unsupervised}. Language specific models alleviate these issues with a custom vocabulary which captures language subtleties\footnote{For example, in Arabic, \cite{antoun-etal-2020-arabert} argue that while the definite article “Al”, which is equivalent to “the” in English, is always prefixed to other words, it is not an intrinsic part of that word. While with a BERT-compatible tokenization tokens will appear twice, once with “Al-” and
once without it, AraBERT first segments the words using Farasa \cite{abdelali2016farasa} and then learns the vocabulary, thereby alleviating the problem.} and large magnitudes of pre-training data scraped from several domains \cite{virtanen2019multilingual, antoun-etal-2020-arabert}. However, building language specific LMs brings us back to where we started, preventing us from leveraging the benefits of multilinguality like zero-shot task transfer \cite{hu2020xtreme}, positive transfer between related languages \cite{pires-etal-2019-multilingual,lauscher-etal-2020-zero} and an ability to handle code-mixed text \cite{pires-etal-2019-multilingual, tsai2019small}. We need an approach that encompasses the best of both worlds, i.e., leverage the capabilities of the powerful language-specific LMs while still being multilingual and enabling positive language transfer.

In this paper, we use knowledge distillation (KD) \cite{hinton2015distilling} to achieve this. In the context of language modeling, KD methods can be broadly classified into two categories: task-specific and task-agnostic. In task-specific distillation, the teacher LM is first fine-tuned for a specific task and is then distilled into a student model which can solve that task. Task-agnostic methods perform distillation on the pre-training objective like masked language modeling (MLM) in order to obtain a task-agnostic student model.  Prior work has either used task-agnostic distillation to compress \emph{single}-language teachers \cite{sanh2019distilbert, sun-etal-2020-mobilebert} or used \emph{task-specific} distillation to combine multiple fine-tuned teachers into a multi-task student \cite{liu-etal-2019-multi-task, clark-etal-2019-bam}. The former prevents positive language transfer while the latter restricts the student's capabilities to the tasks and languages in the fine-tuned teacher LMs (as shown in Figure \ref{fig:distil_types}).


We focus on the problem of merging \emph{multiple} pre-trained LMs into a single multilingual student LM in the \emph{task-agnostic} setting. To the best of our knowledge, this is the first effort of its kind, and makes the following contributions: 

\begin{itemize}
    \item We propose \method{}, a task-agnostic distillation approach to merge \emph{multiple} teacher LMs at the \textit{pre-training} stage, to train a strong multilingual student LM that can then be fine-tuned for \textit{any} task on all languages in the student LM. Our approach is more maintainable (fewer models), compute efficient and teacher-architecture agnostic (since we obtain offline predictions).
    \item We use \method{} to \textbf{i)} combine \emph{monolingual} teacher LMs into a single \emph{multilingual} student LM that is competitive with or outperforms individual teachers, \textbf{ii)} combine \emph{multilingual} teacher LMs, such that the overlapping languages can learn from \emph{multiple} teachers.
    \item 
    Through extensive experiments and analysis, we study the importance of typological similarity in building multilingual models, and the impact of strong teacher LM vocabularies and predictions in our framework.
\end{itemize}

\section{Related Work}
\textbf{Language Model pre-training} has evolved from learning pre-trained word embeddings \cite{mikolov2013distributed} to contextualized word representations \cite{mccann2017learned, peters2018deep, eriguchi2018zero} and to the most recent Transformer-based \cite{vaswani2017attention} LMs \cite{devlin-etal-2019-bert,liu-etal-2019-robust} with state-of-the-art results on various downstream NLP tasks. Most commonly, these LMs are pre-trained with the MLM objective \cite{taylor1953cloze} on large unsupervised corpora and then fine-tuned on labeled data for the task at hand. Concurrently, multilingual LMs \cite{Lample2019xlm, Siddhant2019evaluating, conneau-etal-2020-unsupervised, chung2021rethinking}, trained on massive amounts of multilingual data, have surpassed cross-lingual word embedding spaces \cite{glavavs2019properly, Ruder_2019} to achieve state-of-the-art in cross-lingual transfer. While \citet{pires-etal-2019-multilingual, wu2019beto} highlight their cross-lingual ability, several limitations have been studied. \citet{conneau-etal-2020-unsupervised} highlight the curse of multilinguality. \citet{hu2020xtreme} highlight that even the best multilingual models do not yield satisfactory transfer performance on the XTREME bechmark covering 9 tasks and 40 languages. Importantly, \citet{wu2020all} and \citet{lauscher-etal-2020-zero} observe that these models significantly under-perform for low-resource languages as representation of these languages in the vocabulary and pre-training corpora are severely limited. \\

\noindent \textbf{Language-specific LMs} are becoming increasingly popular as issues with multilingual language models persist. As language identification systems are extended to 1000+ languages \cite{caswell-etal-2020-language}, increasing capacity for a single model to uniformly represent all languages is prohibitive. Often, practitioners prefer to have a model performing well on a subset of languages that their application calls for. To address this, the community continues its efforts in building strong multi-domain language models using linguistic expertise. A few examples of these are AraBERT \cite{antoun-etal-2020-arabert}, CamemBERT \cite{martin-etal-2020-camembert}, and FinBERT \cite{virtanen2019multilingual}.\footnote{\cite{nozza2020mask} maintain an ever-growing list of BERT models \href{https://bertlang.unibocconi.it/}{here}}\\

\noindent \textbf{Knowledge Distillation} in pre-trained LMs has most commonly been used for task-specific model compression of a teacher into a single-task student \cite{tang2019distilling, kaliamoorthi2021distilling}. This has been extended to perform task-specific distillation of multiple single-task teachers into one multi-task student \cite{clark-etal-2019-bam, liu2020mkd, turc2019well}. In the task-agnostic scenario, prior work has focused on distilling a single large teacher model into a student model leveraging teacher predictions \cite{sanh2019distilbert} or internal teacher representations \cite{sun-etal-2020-mobilebert, sun2019patient, wang2020minilm} with the goal of model compression. To the best of our knowledge, this is the first attempt to perform task-agnostic distillation from  \emph{multiple teachers} into a \emph{single task-agnostic student}. In the context of neural machine translation, \citet{tan2019multilingual} come close to our work where they attempt to combine multiple single language-pair teacher models to train a multilingual student. However, our work differs from theirs in three key aspects: 1) our students are task-agnostic while theirs are task-specific, 2) we can leverage pre-existing teachers while they cannot, and 3) we support teachers with overlapping sets of languages while they only consider single language-pairs teachers.
\section{\method{}}
\label{method}
\textbf{Notations}: Let $\mathrm{K}$ denote the set of languages we train our student LM on and $\mathrm{T}$ denote the set of teacher LMs input to \method{}\footnote{Note that $\mathrm{T}$ can comprise of monolingual or multilingual models}. Consequently, $\mathrm{T_k}$ denotes the set of teacher LMs trained on language $\mathrm{k}$, where \( \mathrm{|T_{k}|\geq1} \) $\forall$ $\mathrm{k \in K}$.
\subsection{Workflow}
\label{workflow}
An overview of \method{} is presented in \reffig{fig:architecture}. Here we detail each step involved in training the student LM from multiple teacher LMs.\\ \\
\textbf{Step 1: Input}\\The input to \method{} is a set of pre-trained teacher LMs and pre-training transfer corpora for all the languages we wish to train our student LM on. With reference to \reffig{fig:architecture}, the student LM is trained on $\mathrm{K=}$\{English (en), Spanish (es), Hindi (hi), Korean (ko)\}. We combine four teacher LMs comprising of three monolingual and one multilingual LM. The monolingual LMs are trained on English ($\mathrm{M_{en}}$), Spanish ($\mathrm{M_{es}}$), and Korean ($\mathrm{M_{ko}}$) while the multilingual LM is trained on English and Hindi ($\mathrm{M_{en,hi}}$). Therefore, for each language, the corresponding set of teacher LMs ($\mathrm{T_k}$) can be defined as: [$\mathrm{T_{en}=\{M_{en}, M_{en,hi}\}, T_{es}=\{M_{es}\},}$\\ $\mathrm{T_{hi}=\{M_{en,hi}\}, T_{ko}=\{M_{ko}\}}$]. First, the pre-training transfer corpora is tokenized and masked for each language using their respective teacher LM's tokenizer. For the language with two teachers, English, we tokenize each example using both the teacher LMs. \\

\noindent \textbf{Step 2: Offline Teacher LM Evaluation}\\We now obtain predictions and logits for each masked, tokenized example in each language, by evaluating their respective teacher LMs. For English, we obtain predictions from both $\mathrm{M_{en}}$ and $\mathrm{M_{en,hi}}$ on their respective copies of each training example. In an ideal situation, we believe that multiple strong teachers can present a multi-view generalisation to the student as each teacher learns different features in training. Let $\mathrm{x}$ denote a sequence of tokens where \( \mathrm{x_m} = \{\mathrm{x_1, x_2, x_3 ... x_n} \} \) denote the masked tokens, and $\mathrm{x_{-m}}$ denote the non-masked tokens. Let $\mathrm{v}$ be the vocabulary of student LM $\mathrm{\theta_s}$. In the conventional case of learning from gold labels, we minimize the cross-entropy of student logit distribution for a masked word $\mathrm{x_{m_i}}$, with the \emph{one-hot label} $\mathrm{v_{j}}$, given by:
\vspace{-0.5mm}
\begin{align}
\label{gold_eqn}
\mathrm{P(x_{m_i},v_j)\!=\!\textbf{1}(x_{m_i}\!=\!v_j) \times log\:p(x_{m_i}\!=\!v_j | x_{-m}; \theta_s)}
\end{align} 
With the teacher evaluations, we obtain predictions (and corresponding logits) of the teacher for the masked tokens. Let us denote the teacher output probability distribution (softmax over logits) for token $\mathrm{x_{m_i}}$ by $\mathrm{Q(x_{m_i} | x_{-m}; \theta_t)}$. Therefore, in addition to the loss from gold labels, we minimize the entropy between the student logits and the \emph{teacher distribution}, given by :
\vspace{-0.3mm}
\begin{align}
\mathrm {\hat{P}(x_{m_i},v_j)\!=\!Q(x_{m_i}\!=\!v_j|x_{-m};\theta_{t})} \times \nonumber \\
  \mathrm{log\:p(x_{m_i}\!=\!v_j|x_{-m};\theta_s)}
\label{kd_eqn}
\end{align}

\noindent It is extremely burdensome (both memory and time) to load multiple teacher LMs and obtain predictions during training. Hence, we first store the \emph{top-k} logits for each masked word offline, loading and normalizing them during student LM training, similar to \cite{tan2019multilingual}. Additionally, obtaining offline predictions gives one the freedom to use expensive teacher LMs without increasing the student model training costs and makes our framework teacher-architecture agnostic.\\

\noindent \textbf{Step 3: Vocab Mapping}\\A deterrent in attempting to distill from multiple pre-trained teacher LMs is that each LM has its own vocabulary. This makes it non-trivial to uniformly process an input example for consumption by both the teacher and student LMs. Our student model's vocabulary is the union of \emph{all} teacher LM vocabularies. In the vocab mapping step, the \textit{input indices}, \textit{prediction indices}, and the \textit{gold label indices}, obtained after evaluation from each teacher LM are processed using a teacher$\rightarrow$student vocab map. This converts each teacher token index to its corresponding student token index, ready for consumption by the student model. For simplicity, each teacher and student LM uses WordPiece tokenization~\cite{schuster2012japanese,wu2016googles} in all our experiments. \\

\noindent \textbf{Step 4: Student LM Training}\\The processed \textit{input indices}, \textit{prediction indices}, and \textit{gold label indices} can now be used to train the multilingual student LM. In training, examples from different languages are shuffled together, even within a batch. We train the student LM with the MLM objective. Let $\mathrm{L_{MLM}}$ denote the MLM loss from gold labels. Therefore, with reference to \refeqn{gold_eqn} : \[ \mathrm{L_{MLM} (x_m|x_{-m})= - \frac{1}{n} \sum_{i=1}^{n}\sum_{j=1}^{|v|} P(x_{m_i},v_j)} \] In addition to learning from gold labels, we use teacher predictions as soft labels and minimize the cross entropy between student and teacher distributions. Let $\mathrm{L_{KD}}$ denote the KD loss from a single teacher LM. With reference to \refeqn{kd_eqn}: \[\mathrm{L_{KD} (x_m|x_{-m})= - \frac{1}{n} \sum_{i=1}^{n}\sum_{j=1}^{|v|}\hat{P}(x_{m_i},v_j)} ;\]
The total loss across all languages is minimized, as shown below: \[\mathrm{L_{ALL}=\sum_{k=1}^{K}\lambda(L_{KD}^{T_k}) + (1-\lambda)L_{MLM}^{k}}\]
In the case of multiple teacher LMs, we have $\mathrm{n}$ tokenized instances for a given  example (where $\mathrm{n}$ denotes the number of teachers for a particular language). In this case, each example in English has two copies -- one tokenized using $\mathrm{M_{en}}$ and another using $\mathrm{M_{en,hi}}$. Thus, we explore two possibilities of training in this multi-teacher scenario :
\begin{itemize}
    \item Include all the copies in training. Here the model is exposed to $\mathrm{n}$ different teacher LM predictions, each presenting a multi-view generalisation to the student LM.
    \item Include the best copy in training. The best copy is the one having minimum teacher LM loss for a given example. Here the model is only exposed to the best teacher LM predictions for each example.
\end{itemize}

\section{Experiments}
In this section, we aim to answer the following questions : \\

\noindent \textbf{1)} How effective is \method{} in combining \emph{monolingual} teacher LMs, to train a \emph{multilingual} student LM that leverages the benefits of multilinguality while performing competitively with individual teacher LMs? (\refsec{exp1}) \\

\noindent \textbf{2)} How effective is \method{} in combining \emph{multilingual} teacher LMs, trained on an overlapping set of languages, such that each language can benefit from \emph{multiple} teachers? (\refsec{exp2}) \\

\noindent \textbf{3)} How important are the teacher LM vocabulary and predictions in \method{}? Further, can \method{} enable pre-trained zero-shot transfer? (\refsec{analysis}) \\
\subsection{Setup}
\textbf{Data}: For all our experiments, we use Wikipedia data as pre-training transfer corpora to train the student model, irrespective of the data used in training individual teacher LMs. We use $\mathrm{\alpha=0.7}$ for exponential smoothing of data across languages, similar to mBERT \cite{devlin-etal-2019-bert}.\\

\noindent \textbf{Model Size}: Since transformer-based models perform better as capacity increases \cite{conneau-etal-2020-unsupervised, arivazhagan2019massively}, we keep the number of parameters close to mBERT ($\sim$178M) by appropriately modifying the vocabulary embedding size (like \citet{lan2019albert}) to isolate the positive effects of learning from teacher LMs.\\

\noindent \textbf{Distillation Parameters}: We have two hyper-parameter choices here: 1) k in top-k logits - as it increases, we observe that while performances remain similar, storing k$>$8 number of predictions for each masked word offline significantly increases resource requirements\footnote{More details in \refapp{top_k}}. Hence, we set k=8 in all our experiments. 2) the value of \( \lambda \) in the loss function, which decides the proportion of teacher loss, is annealed through training similar to \citet{clark-etal-2019-bam}.\\

\noindent \textbf{Evaluation Metrics}: We report F1 scores for structured prediction tasks (NER, POS), accuracy (Acc.) scores for sentence classification tasks (XNLI, PAWS-X), and F1/Exact Match (F1/EM) scores for question answering tasks (XQuAD, MLQA, TyDiQA). We also report a task-specific \emph{relative deviation from teachers} (\textbf{RDT}) (in \%) averaged across all languages ($\mathrm{n}$). For each task, \textbf{RDT} is calculated as:
 
\begin{align}
\label{rdt_eqn}
\mathrm{RDT(S, \{T_1, ..., T_n\}) = \frac{100}{n} \sum_{i = 1}^{n} \frac{(P_{T_i} - P_{S})}{P_{T_i}}}
\end{align}

\noindent where $\mathrm{P_{T_i}}$ and $\mathrm{P_{S}}$ are performances of the $\mathrm{i^{th}}$ teacher and student LMs, respectively.

\begin{table}[t]
\label{monolingual_models}
\scalebox{0.55}{
\begin{tabular}{c | c | c | c }
\hline {Student} & {Language} & {Language Family} & {Model} \\ 
\hline
\multirow{4}{*}{$\mathrm{Student_{similar}}$} & {English} & {Indo-European} & {BERT\cite{devlin-etal-2019-bert}} \\
& German & {Indo-European} & {DeepSet\cite{chan2020germans}} \\
& {Italian} & {Indo-European} & {ItalianBERT\cite{stefan_schweter_2020_4263142}}\\
& {Spanish} & {Indo-European} & {BETO\cite{CaneteCFP2020}}\\ \hdashline
\multirow{5}{*}{$\mathrm{Student_{dissimilar}}$} & {Arabic} & {Afroasiatic} & {AraBERT\cite{antoun-etal-2020-arabert}} \\
& {English} & {Indo-European} & {BERT\cite{devlin-etal-2019-bert}} \\
& {Finnish} & {Uralic} & {FinBERT\cite{virtanen2019multilingual}} \\
& {Turkish} & {Turkic} & {BERTurk\cite{stefan_schweter_2020_3770924}}\\
& {Chinese} & {Sino-Tibetan} & {ChineseBERT\cite{devlin-etal-2019-bert}} \\
\hline
\end{tabular}}
\caption{\label{table:mono_models} \emph{Monolingual BERT Models} used as teacher LMs. Please refer to \refsec{exp1} for details.}
\end{table}

\begin{table}[t]
\scalebox{0.75}{
\begin{tabular}{c | c | c c c c c c c c }
\hline \multirow{2}{*}{Language} & \multirow{2}{*}{Model} & {NER} & {UDPOS} & {QA} \\ 
& & \textbf{F1} & \textbf{F1} & \textbf{F1/EM} \\
\hline
\multirow{2}{*}{English} & BERT & 89.5 & 96.6 & 87.1/78.6 \\
& ${\mathrm{Student}_{\mathrm{similar}}}$ & 89.8 & 96.3 & 89.8/82.1 \\
\hline
\multirow{2}{*}{German} & DeepsetBERT & 93.0 & 98.3 & - \\
& ${\mathrm{Student}_{\mathrm{similar}}}$ & 93.9 & 98.3 & - \\
\hline
\multirow{2}{*}{Italian} & ItalianBERT & 94.5 & 98.6 & 73.5/61.6  \\
& ${\mathrm{Student}_{\mathrm{similar}}}$ & 95.2 & 98.6 &  75.8/63.8 \\
\hline
\multirow{2}{*}{Spanish} & BETO & 94.2 & 99.0 & 74.9/56.6 \\
& ${\mathrm{Student}_{\mathrm{similar}}}$ & 94.7 & 98.9 & 76.5/58.4 \\ 
\hdashline
& \multirow{2}{*}{{RDT(\%)}} & \multirow{2}{*}{\textbf{+0.6}} & \multirow{2}{*}{\textbf{-0.1}} & \multirow{2}{*}{\textbf{+2.8/+3.7}} \\
& & & & & \\
\hdashline 
\multirow{2}{*}{Arabic} & AraBERT & {94.3} & 96.3 & {83.1/68.6} \\
& $\mathrm{Student}_{\mathrm{dissimilar}}$ & 93.7 & {96.4} & 81.3/66.6 \\
\hline
\multirow{2}{*}{Chinese} & ChineseBERT & {83.0} & {96.9} & {81.8/81.8} \\
& $\mathrm{Student}_{\mathrm{dissimilar}}$ & 82.6 & 96.8 & 80.8/80.8 \\
\hline
\multirow{2}{*}{English} & BERT & {89.5} & {96.6} & 87.1/78.6 \\
& $\mathrm{Student}_{\mathrm{dissimilar}}$ & {89.5} & 96.3 & {88.6/80.7} \\
\hline
\multirow{2}{*}{Finnish} & FinBERT & {94.4} & 97.9 & {81.0/68.8} \\
& $\mathrm{Student}_{\mathrm{dissimilar}}$ & {94.4} & 95.5 & 77.7/65.9 \\
\hline
\multirow{2}{*}{Turkish} & BERTurk & 95.2 & {95.6} & {76.7/59.8} \\
& $\mathrm{Student}_{\mathrm{dissimilar}}$ & {95.4} & 92.9 & 76.2/59.1 \\
\hdashline
& \multirow{2}{*}{{RDT(\%)}} & \multirow{2}{*}{\textbf{-0.2}} & \multirow{2}{*}{\textbf{-1.1}} & \multirow{2}{*}{\textbf{-1.3/-1.4}} \\
& & & & & \\
\hline 
\end{tabular}}
\caption{\label{monolingual_result} Results for \emph{monolingual teacher LMs and multilingual students} on downstream tasks as described in \refsec{exp1}. Relative deviations of 5\% or less from teacher (i.e., $\mathrm{RDT} \ge -5\%$) are marked in bold. We find that ${\mathrm{Student}_{\mathrm{similar}}}$ outperforms individual teacher LMs, with a maximum gain of upto {+2.8/+3.7\%} for QA, while ${\mathrm{Student}_{\mathrm{dissimilar}}}$ is competitive with teacher LMs, with a maximum drop of {-1.3/-1.4\%} for QA. Please refer to \refsec{exp1} for details.}
\end{table}

\subsection{Monolingual Teacher LMs}
\label{exp1}
\textbf{Pre-training}: In this experiment, we use pre-existing monolingual teacher LMs, as shown in Table \ref{table:mono_models}, to train a multilingual student LM on the union of all teacher languages. In this setup, \( \mathrm{|T_{k}|=1} \) $\forall$ $\mathrm{k \in K}$, i.e., each language can learn from its respective monolingual teacher LM only. 

Our teacher selection and setup follows a two-step process. First, we aim to select languages having pre-trained monolingual LMs available, and evaluation sets across a number of downstream tasks. This makes us choose teacher LMs for : Arabic (\emph{ar}), Chinese (\emph{zh}), English (\emph{en}), Finnish (\emph{fi}), German (\emph{de}), Italian (\emph{it}), Spanish (\emph{es}), and Turkish (\emph{tr}). Second, as previous work has evidenced positive transfer between related languages in a multilingual setup \cite{pires-etal-2019-multilingual, wu2020all}, we further group the chosen teacher LMs based on language families as shown in \reftbl{table:mono_models}, where: \\
\textbf{i) $\boldsymbol{\mathrm{Student_{similar}}}$} is trained on four closely related languages from the Indo-European family -- \emph{de}, \emph{en}, \emph{es} and \emph{it}. \\ 
\textbf{ii) $\boldsymbol{\mathrm{Student}_{\mathrm{dissimilar}}}$} is trained on languages from different language families -- \emph{ar}, \emph{en}, \emph{fi}, \emph{tr} and \emph{zh}. \\

\noindent Both student LMs have a BERT-base architecture. ${\mathrm{Student}_{\mathrm{similar}}}$ has a vocabulary size of 99,112 with a total of 162M parameters, while ${\mathrm{Student}_{\mathrm{dissimilar}}}$ has a vocabulary size of 180,996 with a total of 225M parameters. We keep a batch size of 4096 and train for 250,000 steps with a maximum sequence length of 512. \\ \\
\textbf{Fine-tuning}: We evaluate both the teacher and student LMs on three downstream tasks with in-language fine-tuning for each task\footnote{More details in \refapp{finetuning_details}} : \\ 

\noindent \textbf{i) }\textbf{Named Entity Recognition} \emph{(NER)}: We use the WikiAnn \cite{pan2017cross, rahimi2019massively} dataset for all languages. \\
\textbf{ii) }\textbf{Part-of-Speech Tagging} \emph{(UDPOS)}: We use the Universal Dependencies v2.6 \cite{11234/1-3226} dataset for all languages. \\
\textbf{iii) }\textbf{Question Answering} \emph{(QA)}: We use DRCD for \emph{zh} \cite{shao2018drcd}, TQuAD\footnote{\href{https://tquad.github.io/turkish-nlp-qa-dataset}{https://tquad.github.io/turkish-nlp-qa-dataset}} for \emph{tr}, SQuADv1.1 \cite{rajpurkar2016squad} for \emph{en}, SQuADv1.1-translated for \emph{it} \cite{10.1007/978-3-030-03840-3_29} and \emph{es} \cite{carrino-etal-2020-automatic} and the TyDiQA-GoldP dataset \cite{clark-etal-2020-tydi} for \emph{ar} and \emph{fi}. \\

\begin{table}[t]
\centering
\scalebox{0.65}{
\begin{tabular}{c | c | c | c | c }
\hline \multirow{2}{*}{Student} & \multirow{2}{*}{Language} & {Teacher LM} & {Student LM} & \multirow{2}{*}{\% of Data} \\
& & Tokens & Tokens & \\
\hline
\multirow{6}{*}{$\mathrm{Student_{similar}}$} & {English} & 3300M & 2285M & 69.25\% \\
& German & 23723M & 847M & 3.57\% \\
& Italian & 13139M & 506M & 3.85\% \\
& Spanish & 3000M & 639M & 21.31\% \\ 
\hdashline
& \multirow{2}{*}{\textbf{Total}} & \multirow{2}{*}{\textbf{43162M}} & \multirow{2}{*}{\textbf{4277M}} & \multirow{2}{*}{\textbf{9.9\%}} \\
& & & & \\
\hdashline
\multirow{5}{*}{$\mathrm{Student_{dissimilar}}$} & {Arabic} & 8600M & 135M & 1.58\% \\
& {English} & 3300M & 2285M & 69.25\% \\
& {Finnish} & 3000M & 83M & 2.77\% \\
& {Turkish} & 4405M & 60M & 1.36\% \\
& {Chinese} & 71M & 71M & 100.00\% \\
\hdashline
& \multirow{2}{*}{\textbf{Total}} & \multirow{2}{*}{\textbf{19376M}} & \multirow{2}{*}{\textbf{2634M}} & \multirow{2}{*}{\textbf{13.6\%}} \\
& & & & \\
\hline
\end{tabular}}
\caption{\label{data_exp1} \emph{Number of Tokens (in Millions)} in the teacher (\reftbl{table:mono_models}) and student LMs as described in \refsec{exp1}}
\end{table}

\begin{table*}[t]
\centering
\scalebox{0.65}{
\begin{tabular}{c | l | l | c c c c c c c c }
\hline
\multirow{2}{*}{Languages} & \multirow{2}{*}{Model} & \multirow{2}{*}{Teacher} & {PANX} & {UDPOS} & {PAWSX} & {XNLI} & {XQUAD} & {MLQA} & {TyDiQA} & \multirow{2}{*}{\textbf{Avg.}}\\ 
& & & \textbf{F1} & \textbf{F1} & \textbf{Acc.} &\textbf{Acc.} & \textbf{F1/EM} & \textbf{F1/EM} & \textbf{F1/EM} \\
\hline
\multirow{7}{*}{MuRIL Languages} & $\mathrm{mBERT}$ & - & 58.8 & 68.5 & 93.4 & 66.2 & 70.3/57.5 & 65.0/50.8 & 62.5/52. & 69.2 \\
& $\mathrm{MuRIL}$ & - & {76.9} & {74.5} & 95.0 & {74.4} & {77.7/64.2} & {73.6/58.6} & {76.1/60.2} & {78.3} \\
& $\mathrm{Student}_{\mathrm{MuRIL}}$ & $\mathrm{MuRIL}$ & 69.3 & 72.3 & {95.4} & 71.9 & 75.7/62.1 & 72.0/56.3 & 70.7/59.2 & 75.3\\
& $\mathrm{Student}_{\mathrm{mBERT}}$ & $\mathrm{mBERT}$ & 38.1 & 52.1 & 93.5 & 64.8 & 56.9/44.8 & 51.1/39.7 & 41.6/33.9 & 56.9\\
& $\mathrm{Student}_{\mathrm{Both\_all}}$ & $\mathrm{mBERT}$ + $\mathrm{MuRIL}$ & 67.9 & 72.3 & 94.5 & 71.1 & 76.1/62.9 & 70.4/55.5 & 70.8/55.3 & 74.7 \\
& $\mathrm{Student}_{\mathrm{Both\_best}}$ & $\mathrm{mBERT}$ + $\mathrm{MuRIL}$ & 68.5 & 71.5 & 93.9 & 70.7 & 77.7/64.3 & 70.8/55.6 & 70.6/58.4 & 74.8 \\
\cdashline{2-11} 
\vspace{0.2mm}
& \multicolumn{2}{|l|}{\multirow{2}{*}{$\mathrm{RDT}(\mathrm{Student}_{\mathrm{MuRIL}}, \mathrm{mBERT})$ (\%)}}  & \multirow{2}{*}{\textbf{+17.9}} & \multirow{2}{*}{\textbf{+5.6}} & \multirow{2}{*}{\textbf{+2.1}} & \multirow{2}{*}{\textbf{+8.6}} & \multirow{2}{*}{\textbf{+7.7/+8}} & \multirow{2}{*}{\textbf{+10.8/+10.8}} & \multirow{2}{*}{\textbf{+13.1/+12.3}} & \multirow{2}{*}{\textbf{+8.8}}\\
& \multicolumn{2}{|l|}{} & & & & & & & & \\
\cdashline{2-11}
& \multicolumn{2}{|l|}{\multirow{2}{*}{$\mathrm{RDT}(\mathrm{Student}_{\mathrm{MuRIL}}, \mathrm{MuRIL})$ (\%)}} & \multirow{2}{*}{-9.9} & \multirow{2}{*}{\textbf{-3}} & \multirow{2}{*}{\textbf{+0.4}} & \multirow{2}{*}{\textbf{-3.4}} & \multirow{2}{*}{\textbf{-2.6/-3.3}} & \multirow{2}{*}{\textbf{-2.2/-3.9}} & \multirow{2}{*}{-7.1/\textbf{-1.7}} & \multirow{2}{*}{\textbf{-3.8}} \\
& \multicolumn{2}{|l|}{} & & & & & & & & \\
\hline
\multirow{6}{*}{Non MuRIL Languages} & $\mathrm{mBERT}$ & - & 63.5 & 71.1 & 80.2 & 65.9 & 62.2/47.1 & 59.7/41.4 & 60.4/46.1 & 66.1\\
& $\mathrm{Student}_{\mathrm{MuRIL}}$ & $\mathrm{mBERT}$ & 63.9 & 72.8 & 83.3 & 68.7 & {66.5/51.2} & {63.1/44.4} & {61.7/45.0} & {68.6}\\
& $\mathrm{Student}_{\mathrm{mBERT}}$ & $\mathrm{mBERT}$ & {64.6} & 72.1 & {84.0} & {68.8} & 64.5/49.0 & 61.1/42.7 & 58.9/44.1 & 67.7 \\
& $\mathrm{Student}_{\mathrm{Both\_all}}$ & $\mathrm{mBERT}$ & 64.1 & 72.6 & 83.9 & 68.1 & 61.3/47.1 & 60.5/42.2 & 59.7/44.0 & 67.2 \\
& $\mathrm{Student}_{\mathrm{Both\_best}}$ & $\mathrm{mBERT}$ & 63.3 & 72.6 & 83.2 & 67.2 & 66.0/50.6 & 61.4/43.2 & 62.4/46.5 & 68.0 \\
\cdashline{2-11}
& \multicolumn{2}{|l|}{\multirow{2}{*}{$\mathrm{RDT}(\mathrm{Student}_{\mathrm{MuRIL}}, \mathrm{mBERT})$ (\%)}} & \multirow{2}{*}{\textbf{+0.6}} & \multirow{2}{*}{\textbf{+2.4}} & \multirow{2}{*}{\textbf{+3.9}} & \multirow{2}{*}{\textbf{+4.3}} & \multirow{2}{*}{\textbf{+6.9/+8.7}} & \multirow{2}{*}{\textbf{+5.7/+7.2}} & \multirow{2}{*}{\textbf{+2.2/-2.4}} & \multirow{2}{*}{\textbf{+3.8}}\\
& \multicolumn{2}{|l|}{} & & & & & & & & \\
\hline
\end{tabular}}
\caption{\label{table_exp2} \emph{Results for multilingual teacher and student LMs} on the XTREME benchmark. We compare performances of three student LM variants as described in \refsec{exp2} to the two teachers mBERT and MuRIL. Relative deviations of 5\% or less from teacher (i.e., $\mathrm{RDT} \ge -5\%$) are marked in bold. Overall, we find that $\mathrm{Student}_{\mathrm{MuRIL}}$ performs the best among all student variants and report its $\mathrm{RDT}$ (in \%) (\refeqn{rdt_eqn}) from the two teachers. Please refer to \refsec{exp2} for a detailed analysis.}
\end{table*}

\noindent \textbf{Results}: We report results of our teacher and student LMs in \reftbl{monolingual_result}. Overall, we find that ${\mathrm{Student}_{\mathrm{similar}}}$ outperforms individual teacher models on NER (+0.6\%) and QA (+2.8/3.7\%) while performing competitively on UDPOS (-0.1\%). ${\mathrm{Student}_{\mathrm{dissimilar}}}$ is competitive with the teacher LMs with only small differences of up to 1.3/1.4\% (QA), as shown in \reftbl{monolingual_result}. For each language, we find ${\mathrm{Student}_{\mathrm{similar}}}$ is either competitive or outperforms its respective teacher LM. Our results provide evidence for positive transfer across languages in two ways. First, we observe that ${\mathrm{Student}_{\mathrm{similar}}}$ outperforms ${\mathrm{Student}_{\mathrm{dissimilar}}}$ for the common language - English. Given that the English teacher (BERT) and the pre-training transfer corpora\footnote{In fact, we can hypothesize that $\mathrm{Student}_{\mathrm{dissimilar}}$ sees more English tokens as compared to ${\mathrm{Student}_{\mathrm{similar}}}$ because the Non-English languages in ${\mathrm{Student}_{\mathrm{dissimilar}}}$ are relatively low resourced (a sum total of 349M unique tokens) in comparison to ${\mathrm{Student}_{\mathrm{similar}}}$ (a sum total of 1992M unique tokens) as shown in \reftbl{data_exp1}} is common for both student LMs, we can attribute this gain to the fact that English is trained with linguistically and typologically similar languages in ${\mathrm{Student}_{\mathrm{similar}}}$. Second, ${\mathrm{Student}_{\mathrm{similar}}}$ outperforms its teacher LMs while $\mathrm{Student}_{\mathrm{dissimilar}}$ is competitive for \emph{all} languages. These two results across \emph{all} languages point towards ${\mathrm{Student}_{\mathrm{similar}}}$ benefiting from a positive transfer across similar languages. In \reftbl{data_exp1}, we observe that ${\mathrm{Student}_{\mathrm{similar}}}$ is trained on 9.9\% of the total unique tokens seen by its respective teacher LMs and ${\mathrm{Student}_{\mathrm{dissimilar}}}$ lies close with 13.6\%.
Despite this huge disparity in pre-training corpora, student LMs are competitive with their teachers. This encouraging result proves that even with very limited data, \method{} enables one to combine strong monolingual teacher LMs to train competitive student LMs that can leverage the benefits of multilinguality.

%

\begin{table*}[t]
\centering
\scalebox{0.7}{
\begin{tabular}{c | c | c | c c c c c c c c }
\hline {Model} & {Vocabulary} & {Labels} & {PANX} & {UDPOS} & {PAWSX} & {XNLI} & {XQUAD} & {MLQA} & {TyDiQA} & {Avg.}\\
\hline
$\mathrm{SM1}$ & mBERT & Gold & 63.2 & 73.0 & 94.8 & 71.2 & 70.2/57.9 & 65.1/51.3 & 60.8/48.7 & 71.2\\
$\mathrm{SM2}$ & mBERT$\cup$MuRIL & Gold & \textbf{69.3} & \textbf{73.9} & 95.3 & 71.2 & \textbf{76.2/63.1} & 71.1/56.0 & 70.9/56.0 & \textbf{75.4} \\
{$\mathrm{SM3}$} & mBERT$\cup$MuRIL & Gold+Teacher & \textbf{69.3} & 72.3 & \textbf{95.4} & \textbf{71.9} & 75.7/62.1 & \textbf{72.0/56.3} & \textbf{70.7/59.2} & 75.3 \\
\cdashline{1-11}
$\mathrm{SM2\_100k}$ & mBERT$\cup$MuRIL & Gold & 65.5 & 72.3 & 94.3 & 67.5 & 72.3/58.2 & 66.9/51.5 & 62.5/51.9 &  71.6 \\
{$\mathrm{SM3\_100k}$} & mBERT$\cup$MuRIL & Gold+Teacher & 71.2 & 73.5 & 93.1 & 69.6 & 76.4/62.9 & 69.1/53.9 & 68.6/54.9 & 74.5 \\
\hline
\end{tabular}}
\caption{\label{vocab_results} \emph{Importance of teacher vocabulary and predictions in \method{}}. We observe maximum performance gains, by changing the vocabulary from mBERT in $\mathrm{SM1}$ to (mBERT$\cup$MuRIL) vocabulary in $\mathrm{SM2}$. Here, SM3 is the standard $\mathrm{Student}_\mathrm{MuRIL}$. We also observe that $\mathrm{SM3\_100k}$, trained for 20\% of the total training steps, is competitive to $\mathrm{SM3}$ and significantly outperforms $\mathrm{SM2\_100k}$, highlighting the importance of teacher LM predictions in a limited data scenario. Please see \refsec{analysis} for details. }
\end{table*}

\subsection{Multilingual Teacher LMs}
\label{exp2}
\textbf{Pre-training}: In this experiment, we make use of pre-existing multilingual models: mBERT and MuRIL. mBERT is trained on 104 languages and MuRIL covers 12 of these (11 Indian languages + English): Bengali (\emph{bn}), English (\emph{en}), Gujarati (\emph{gu}), Hindi (\emph{hi}), Kannada (\emph{kn}), Malayalam (\emph{ml}), Marathi (\emph{mr}), Nepali (\emph{ne}), Punjabi (\emph{pa}), Tamil (\emph{ta}), Telugu (\emph{te}), and Urdu (\emph{ur}), with higher performance for these languages on the XTREME benchmark. We train the student model on all 104 languages. In this case, the \emph{MuRIL Languages} (MuL) have two teachers (mBERT and MuRIL) and the \emph{Non-MuRIL Languages} (Non-MuL) can learn from mBERT only. Therefore, while we only use mBERT as the teacher LM for Non-MuL across all experiments, we consider three possibilities for MuL : \\ \\
\textbf{i)} $\boldsymbol{\mathrm{Student}_{\mathrm{MuRIL}}}$: We only use MuRIL as the teacher LM and each input training example is tokenized using MuRIL. \\
\textbf{ii)} $\boldsymbol{\mathrm{Student}_{\mathrm{mBERT}}}$: We only use mBERT as the teacher LM and each input training example is tokenized using mBERT. \\
\textbf{iii)} $\boldsymbol{\mathrm{Student}_{\mathrm{Both}}}$: As highlighted in \refsec{method}, we consider two possibilities to incorporate both teacher LM predictions in training:
\begin{itemize}
    \item $\boldsymbol{\mathrm{Student}_{\mathrm{Both\_all}}}$: Tokenize each input example using mBERT and MuRIL separately and include both copies in training.
    \item $\boldsymbol{\mathrm{Student}_{\mathrm{Both\_best}}}$: Tokenize each input example using mBERT and MuRIL separately and include only the best copy in training. The best copy is the one having minimum teacher LM loss for the example.
\end{itemize}

\noindent Note, it is non-trivial to tokenize each example in a way that is compatible with all teacher LMs. One must resort to tokenization using an intersection of vocabularies which is sub-optimal. \\

\noindent All the student LMs use a BERT-base architecture and have a vocabulary size of 288,973. We reduce our embedding dimension to 256 as opposed to 768 to bring down the model size to be around 160M, comparable to mBERT (178M). We keep a batch size of 4096 and train for 500,000 steps with a maximum sequence length of 512. \\ 

\noindent \textbf{Finetuning}: We report zero-shot performance for all languages in the XTREME \cite{hu2020xtreme} benchmark\footnote{More details in \refapp{finetuning_details}}. \\

\noindent \textbf{Results}: We report results of our teacher and student LMs in \reftbl{table_exp2}. Overall, we find that $\mathrm{Student}_{\mathrm{MuRIL}}$ performs the best among all student variants. For Non-MuL, $\mathrm{Student}_{\mathrm{MuRIL}}$ beats the teacher (mBERT) by an average relative score of 3.8\%. For MuL, $\mathrm{Student}_{\mathrm{MuRIL}}$ beats one teacher (mBERT) by 8.8\%, but underperforms the other teacher (MuRIL) by 3.8\%. There can be two factors at play here. MuRIL is trained on monolingual and parallel data \footnote{More details in \refapp{pretraining_details}} while the student LMs only see $\sim$22\% of unique tokens in comparison. MuRIL also has different language sampling strategies ($\mathrm{\alpha=0.3}$ as opposed to $\mathrm{0.7}$ in our setting, where a lower $\mathrm{\alpha}$ value upsamples more rigorously from the tail languages), which have a significant role to play in multilingual model performances \cite{conneau-etal-2020-unsupervised}. We also observe a significant drop in $\mathrm{Student}_{\mathrm{mBERT}}$'s performance for MuL when compared to the other student LM variants. This might be because the input is tokenized using the mBERT tokenizer which prevents learning from MuRIL tokens in the student vocabulary.
For $\mathrm{Student}_{\mathrm{Both}}$, we do not observe much of a difference between $\mathrm{Student}_{\mathrm{Both\_all}}$ and $\mathrm{Student}_{\mathrm{Both\_best}}$. This observation may differ with one's choice of teacher LMs depending on how well it performs for a particular language. In our case, we don't observe much of a difference in incorporating mBERT predictions for MuL. 

\subsection{Further Analysis}
\label{analysis}

\textbf{The importance of vocabulary and teacher LM preditions}: In \reftbl{table_exp2}, we see that $\mathrm{Student}_{\mathrm{MuRIL}}$ significantly outperforms mBERT for MuL, despite both being trained on Wikipedia corpora, and having comparable model sizes. With regard to MuL, $\mathrm{Student}_{\mathrm{MuRIL}}$ differs from mBERT in two main aspects -- \textbf{i)} $\mathrm{Student}_{\mathrm{MuRIL}}$'s vocabulary is a union of mBERT and MuRIL vocabularies. \textbf{ii)} $\mathrm{Student}_{\mathrm{MuRIL}}$ is trained with additional MuRIL predictions as soft labels. To disentangle the role both these factors play in $\mathrm{Student}_{\mathrm{MuRIL}}$'s improved performance, we train two models : \\
\noindent \textbf{i)} $\mathrm{SM1}$ is trained exactly like $\mathrm{Student}_{\mathrm{MuRIL}}$, but with mBERT vocabulary and on gold labels. \\
\textbf{ii)} $\mathrm{SM2}$ is trained using $\mathrm{Student}_{\mathrm{MuRIL}}$'s vocabulary (mBERT $\cup$ MuRIL) but on gold labels only, without teacher predictions. \\

\noindent The results are summarized in \reftbl{vocab_results}. Note, we refer to $\mathrm{Student}_{\mathrm{MuRIL}}$ as $\mathrm{SM3}$. Overall, we observe a \textbf{$\sim$4.2\%} gain in average performance for $\mathrm{SM2}$ over $\mathrm{SM1}$. This clearly highlights that given fixed data and model capacity, LM training significantly benefits by incorporating a strong teacher's vocabulary. \\

\noindent Furthermore, we also observe that $\mathrm{SM2}$ and $\mathrm{SM3}$ achieve competitive performances despite $\mathrm{SM3}$ being additionally trained on teacher LM labels. To motivate the need for teacher predictions, \citet{hinton2015distilling} argue that when soft targets have high entropy, they provide much more information per training case than hard targets and can be trained on \emph{much less data} than the original cumbersome model. In our case, we hypothesize that training on 500,000 steps exposes the model to sufficient data for it to generalize well enough and mask the benefits of teacher LM predictions. To validate this, we evaluate the performances of $\mathrm{SM2}$ and $\mathrm{SM3}$, $\mathrm{20\%}$ into training (i.e. 100,000 steps / 500,000 total steps) as shown in \reftbl{vocab_results}. We observe a \textbf{$\sim$2.9\%} gain in average performance for $\mathrm{SM3}$ over $\mathrm{SM2}$, clearly highlighting the importance of teacher LM predictions in a limited data scenario. This is especially important when one has access to very limited monolingual data and a strong teacher LM for a particular language. \\ 

\noindent \textbf{Pre-trained zero-shot transfer: }Interestingly, $\mathrm{Student}_{\mathrm{MuRIL}}$ performs the best on almost all tasks for \emph{Non-MuL}. This hints at positive transfer from strong teachers to languages that the teacher does not cover at all, due to the shared multilingual representations.\footnote{For example, if you want to train a multilingual model covering English and a closely related low-resource language for which there exists no strong teacher, it may be possible to improve performance for the low resource language using teacher predictions for English only, due to a shared embedding space and possibly shared sub-words.} This would mean that learning from strong teachers can improve the student model's performance in a zero-shot manner on related languages not covered by the teacher. This would make \method{} highly beneficial for low-resource languages that do not have a strong teacher or limited gold data. We leave this exploration to future work.
\section{Conclusion}
In this paper we address the problem of merging multiple pre-trained teacher LMs into a single  multilingual student LM by proposing \method{}, a task-agnostic distillation method. To the best of our knowledge, this is the first attempt of its kind. The student LM learned by \method{} may be further fine-tuned for any task across all of the languages covered by the teacher LMs. Our approach results in better maintainability (fewer models) and is compute efficient (due to offline predictions). We use \method{} to \textbf{i)} combine \emph{monolingual} teacher LMs into one student multilingual LM which is competitive with the teachers, thereby demonstrating positive cross-lingual transfer, and  \textbf{ii)} combine multilingual LMs to train student LMs that learn from \emph{multiple} teachers. Through  experiments on multiple benchmark datasets, we show that student LMs learned by \method{} perform competitively or even outperform teacher LMs trained on orders of magnitude more data. We disentangle the positive impact of incorporating strong teacher LM vocabularies and learning from teacher LM predictions, highlighting the importance of the latter in a limited data scenario. We also find that \method{} enables positive transfer from strong teachers to languages not covered by them (i.e. zero-shot transfer). Our work bridges the gap between the universe of language-specific models and massively multilingual LMs, incorporating benefits of both into one framework.

\section{Acknowledgements}
We would like to thank the anonymous reviewers for their insightful and constructive feedback. We thank Iulia Turc, Ming-Wei Chang, and Slav Petrov for valuable comments on an earlier version of this paper.


\bibliographystyle{acl_natbib}
\bibliography{anthology,acl2021}

\appendix
\section{Appendix}
\subsection{Knowledge Distillation}
\label{background}
We train our LMs with the MLM objective. Let $\mathrm{x}$ denote a sequence of tokens where \( \mathrm{x_m} = \{\mathrm{x_1, x_2, x_3 ... x_n} \} \) denote the masked tokens, and $\mathrm{x_{-m}}$ denote the non-masked tokens. Let $\mathrm{v}$ be the vocabulary of LM \( \mathrm{\theta} \). The log-likelihood loss (cross-entropy with one-hot label) can be formulated as follows:  \[ \mathrm{L_{MLM} (x_m|x_{-m})= - \frac{1}{n} \sum_{i=1}^{n}\sum_{k=1}^{|v|} P(x_{m_i},k)} ; \]
\[\mathrm {P(x_{m_i},k) = \textbf{1}(x_{m_i}=k) log p(x_{m_i}=k | x_{-m}; \theta)} \] \\
In a distillation setup, the student is trained to not only match the one-hot labels for masked words, but also the probability output distribution of the teacher $\mathrm{t}$. Let us denote the teacher output probability distribution for token $\mathrm{x_{m_i}}$ by $\mathrm{Q(x_{m_i} | x_{-m}; \theta_t)}$. The cross entropy between the teacher and student distributions then serves as the distillation loss :
\[\mathrm{L_{KD} (x_m|x_{-m})= - \frac{1}{n} \sum_{i=1}^{n}\sum_{k=1}^{|v|}\hat{P}(x_{m_i},k)} ;\]
\begin{align*}
\mathrm {\hat{P}(x_{m_i},k)= Q(x_{m_i}=k | x_{-m};\theta_{t})} \\
    \mathrm{log p(x_{m_i}=k | x_{-m}; \theta)}   
\end{align*}

The total loss is then defined as : \[ \mathrm{L_{ALL} = \lambda L_{KD} + (1-\lambda) L_{MLM}} \]
With the addition of the teacher, the target distribution is no longer a single one-hot label,  but a smoother distribution with multiple words having non-zero probabilities
which yields in a smaller variance in gradients \cite{hinton2015distilling}. Intuitively, a single masked word can have several valid predictions, which appropriately fit the context.
\subsection{Pre-training Details}
\label{pretraining_details}
\subsubsection{Monolingual Teacher LMs}
We pre-train our student models using the BERT base architecture. ${\mathrm{Student}_{\mathrm{similar}}}$ has a vocabulary size of 99112 and a model size of 162M parameters. $\mathrm{Student}_{\mathrm{different}}$ has a vocabulary size of 180996 and a model size of 225M parameters. We keep a batch size of 4096 and train for 250k steps with a maximum sequence length of 512. We use TPUs, and it takes around 1.5 days to pre-train each student LM.
\subsubsection{Multilingual Teacher LMs}
We pre-train our student models using the BERT base architecture. All student LMs have a vocabulary size of 288973. Hence, we reduce our embedding dimension to 256 as opposed to 768 to bring down the model size to be around 160M, comparable to mBERT (178M). We keep a batch size of 4096 and train for 500k steps with a maximum sequence length of 512. We use TPUs, and it takes around 3 days to pre-train each student LM.\\ \\
We present pre-training data statistics for MuRIL and the student LMs in \reftbl{data_exp2}. Here we only include the monolingual data statistics, but MuRIL is additionally trained on parallel translated and transliterated data.

\begin{table}[t]
\centering
\scalebox{0.7}{
\begin{tabular}{c| c | c | c | c }
\hline \multirow{2}{*}{Teacher} & \multirow{2}{*}{Language} & {Teacher LM} & {Student LM} & \multirow{2}{*}{\% of Data} \\
& & Tokens & Tokens & \\
\hline
\multirow{14}{*}{MuRIL} & Bengali & 1181M & 27M & 2.30\% \\
& English & 6986M & 2816M & 40.30\%  \\
& Gujarati & 173M & 7M & 3.90\% \\
& Hindi & 2368M & 38M & 1.61\% \\
& Kannada & 196M & 15M & 7.64\% \\
& Malayalam & 337M & 14M & 4.17\% \\
& Marathi & 274M & 8M & 3.02\% \\
& Nepali & 231M & 5M & 2.16\% \\
& Punjabi & 141M & 9M & 6.45\% \\
& Tamil & 769M & 26M & 3.34\% \\
& Telugu & 331M & 30M & 8.99\% \\
& Urdu & 722M & 23M & 3.21\% \\
\hdashline
& \multirow{2}{*}{\textbf{Total}} & \multirow{2}{*}{\textbf{13709M}} & \multirow{2}{*}{\textbf{3018M}} & \multirow{2}{*}{\textbf{22\%}} \\
& & & & \\
\hline
\end{tabular}}
\caption{\label{data_exp2} \emph{Number of Tokens (in Millions)} in the teacher (MuRIL) and student LMs as described in \refsec{exp2}. Note, we only show the \emph{MuRIL Languages} here because for \emph{Non-MuRIL Languages}, the teacher (mBERT) and student variants are trained on the same data.}
\end{table}

\begin{table}[t]
\centering
\scalebox{0.75}{
\begin{tabular}{c | c | c | c | c | c }
\hline \multirow{2}{*}{Task} & \multirow{2}{*}{Batch} & Learning & No. of & Warmup & Max. seq. \\ 
& & Rate & Epochs & Ratio & Length \\
\hline
NER & 32 & 3e-5 & 10 & 0.1 & 256 \\
POS & 32 & 3e-5 & 10 & 0.1 & 256\\
QA & 32 & 3e-5 & 10 & 0.1 & 384 \\
\hline 
\end{tabular}}
\caption{\label{hyperparams_exp1} Hyperparameter Details for each fine-tuning task in \refsec{exp1}}
\end{table}

\begin{table*}[t]
\centering
\scalebox{0.8}{
\begin{tabular}{c | c | c c c c c c c c }
\hline {Languages} & {Model} & {PANX} & {UDPOS} & {PAWSX} & {XNLI} & {XQUAD} & {MLQA} & {TyDiQA} \\ 
\hline
\multirow{5}{*}{MuRIL Languages} & mBERT & 58.8 & 68.5 & 93.4 & 66.2 & 70.3/57.5 & 65.0/50.8 & 62.5/52.7 \\
& MuRIL & \textbf{76.9} & 74.5 & 95.0 & \textbf{74.4} & \textbf{77.7/64.2} & \textbf{73.6/58.6} & \textbf{76.1/60.2} \\
& k=8 & 69.3 & 72.3 & \textbf{95.4} & 71.9 & 75.7/62.1 & 72.0/56.3 & 70.7/59.2 \\
& k=128 & 67.5 & 72.8 & 94.4 & 70.7 & 75.5/61.9 & 71.1/56.1 & 70.2/55.4 \\
& k=512 & 69.2 & \textbf{77.2} & 94.7 & 71.3 & 75.6/61.8 & 72.3/56.9 & 68.5/53.9 \\
\hline
\multirow{4}{*}{Non MuRIL Languages} & mBERT & 63.5 & 71.1 & 80.2 & 65.9 & 62.2/47.1 & 59.7/41.4 & 60.4/46.1 \\
& k=8 & 63.9 & 72.8 & 83.3 & \textbf{68.7} & \textbf{66.5/51.2} & 63.1/44.4 & \textbf{61.7/45.0} \\
& k=128 & 63.7 & 72.8 & \textbf{83.4} & 67.9 & 66.1/51.1 & 61.4/43.4 & 62.6/46.7 \\
& k=512 & \textbf{64.8} & \textbf{73.3} & 82.7 & 67.4 & 65.7/50.7 & \textbf{63.6/44.9} & 58.7/44.8 \\
\hline
\multirow{4}{*}{All Languages} & mBERT & 62.5 & 70.6 & 82.0 & 65.9 & 63.7/49.0 & 61.2/44.1 & 61.1/48.3 \\
& k=8 & 65.0 & 72.7 & \textbf{85.0} & \textbf{69.3} & \textbf{68.2/53.2} & 65.6/47.8 & \textbf{64.7/49.7} \\
& k=128 & 64.5 & 72.8 & \textbf{85.0} & 68.4 & 67.9/53.0 & 64.2/47.1 & 65.2/49.6 \\
& k=512 & \textbf{65.7} & \textbf{74.0} & 84.4 & 68.2 & 67.5/52.8 & \textbf{66.1/48.3} & 62.0/47.8 \\
\hline
\end{tabular}}
\caption{\label{table:top_k} \emph{Results of the best performing student model $\mathrm{Student}_{\mathrm{MuRIL}}$} for different top-k values}
\end{table*}

\begin{table}[t]
\scalebox{0.8}{
\begin{tabular}{c | c | c }
\hline {Language} & {Dataset} & {Examples (Train/Dev/Test)} \\ 
\hline
Arabic & AR\_PADT &  6075/909/680\\
Chinese & ZH\_GSD &   3997/500/500\\
English & EN\_EWT &  12543/2002/2077\\
German & DE\_HDT &  15305/18434/18459\\
Finnish & FI\_FTB & 14981/1875/1867 \\
Italian & IT\_ISDT & 13121/564/482 \\
Spanish & ES\_ANCORA & 14305/1654/1721 \\
Turkish & TR\_IMST & 3664/988/983 \\
\hline
\end{tabular}}
\caption{\label{pos} \emph{Universal Dependencies v2.6} overview for each language, used in \refsec{exp1} }
\end{table}

\begin{table}[t]
\scalebox{0.8}{
\begin{tabular}{c | c | c }
\hline {Language} & {Dataset} & {Examples (Train/Test)} \\ 
\hline
Arabic & TyDiQA-GoldP &  14805/921\\
Chinese & DRCD &  26936/3524\\
English & SQuADv1.1 &  87599/10570\\
German & - &  -\\
Finnish & TyDiQA-GoldP &  6855/782\\
Italian &  SQuADv1.1-translated & 87599/10570 \\
Spanish &  SQuADv1.1-translated & 87595/10570\\
Turkish & TQuAD &  8308/892\\
\hline
\end{tabular}}
\caption{\label{qa} \emph{Question Answering datasets}, used in \refsec{exp1} }
\end{table}

\subsection{Fine-tuning Details}
\label{finetuning_details}
\subsubsection{Monolingual Teacher LMs}
\textbf{Data Statistics} We evaluate our monolingual teacher LMs and multilingual student LMs, as described in \refsec{exp1}, on three tasks as follows: \\ \\
\textbf{i) Named Entity Recognition} (NER): We use the WikiAnn \cite{pan2017cross, rahimi2019massively} dataset for all languages. Each language comprises of a train/dev/test split of 20000/10000/10000 tokens. Specifically, we use the huggingface re-packaged implementation of the dataset\footnote{\href{https://huggingface.co/datasets/wikiann}{https://huggingface.co/datasets/wikiann}}. \\ \\
\textbf{ii) Part-of-Speech tagging} (POS): We use the Universal Dependencies v2.6 \cite{11234/1-3226} dataset for all languages. Detailed statistics for each language can be found in \reftbl{pos}. Specifically, we use the huggingface re-packaged implementation of the dataset\footnote{\href{https://huggingface.co/datasets/universal_dependencies}{https://huggingface.co/datasets/universal\_dependencies}}. \\ \\
\textbf{iii) Question Answering} (QA): We use the TyDiQA dataset \cite{clark-etal-2020-tydi} for \emph{ar} and \emph{fi}, SQuADv1.1 \cite{rajpurkar2016squad} for \emph{en}, SQuAD-translated for \emph{it} \cite{10.1007/978-3-030-03840-3_29} and \emph{es} \cite{carrino-etal-2020-automatic}, DRCD for \emph{zh} \cite{shao2018drcd} and TQuAD\footnote{\href{https://tquad.github.io/turkish-nlp-qa-dataset}{https://tquad.github.io/turkish-nlp-qa-dataset}} for \emph{tr}. Detailed statistics for each language can be found in \reftbl{qa}. Note, we use the dev set as our test sets, since most datasets only have a train/dev split. We use 10\% of randomly shuffled training examples as our dev sets. \\

\begin{table}[t]
\centering
\scalebox{0.75}{
\begin{tabular}{c | c | c | c | c | c }
\hline \multirow{2}{*}{Task} & \multirow{2}{*}{Batch} & Learning & No. of & Warmup & Max. seq. \\ 
& & Rate & Epochs & Ratio & Length \\
\hline
PANX & 32 & 2e-5 & 10 & 0.1 & 128 \\
UDPOS & 64 & 5e-6 & 10 & 0.1 & 128\\
PAWSX & 32 & 2e-5 & 5 & 0.1 & 128 \\
XNLI & 32 & 2e-5 & 3 & 0.1 & 128 \\
XQuAD & 32 & 3e-5 & 2 & 0.1 & 384 \\
MLQA & 32 & 3e-5 & 2 & 0.1 & 384 \\
TyDiQA & 32 & 3e-5 & 2 & 0.1 & 384 \\
\hline 
\end{tabular}}
\caption{\label{hyperparams_exp2} Hyperparameter Details for each task in XTREME}
\end{table}

\textbf{Hyperparameter Details}: We use the same hyperparameters for fine-tuning all teacher and student LMs, as shown in \reftbl{hyperparams_exp1}. We report results on the best-performing checkpoint for the validation set. 
\subsubsection{Multilingual Teacher LMs}
\textbf{Data Statistics} We evaluate all the teacher (mBERT and MuRIL) and student ($\mathrm{Student}_{\mathrm{MuRIL}}$, $\mathrm{Student}_{\mathrm{mBERT}}$ and $\mathrm{Student}_{\mathrm{Both}}$) LMs on the XTREME \cite{hu2020xtreme} benchmark. We fine-tune the pre-trained models on English training data for the particular task, except TyDiQA, where we use additional SQuAD v1.1 English training data, similar to \cite{fang2020filter}. All results are computed in a zero-shot setting. \\  \\
\textbf{Hyperparameter Details} We use the same hyperparameters for fine-tuning all teacher and student LMs, as shown in \reftbl{hyperparams_exp2}. We report results on the best-performing checkpoint for the eval set.

\subsection{Different top-k values}
\label{top_k}
We present results for $\mathrm{Student_{MuRIL}}$ trained with different top-k values from teacher predictions in \reftbl{table:top_k}. We observe that while performances remain similar for higher values of k, storage becomes increasingly expensive. Hence, we stick to a value of k=8 in all our experiments.

\end{document}